\pdfoutput=1

\documentclass[11pt]{article}

\usepackage{acl}
\usepackage{expex}
\usepackage{array} 

\lingset{everygla={}, glspace=.3em, glhangstyle=none, glhangindent=0pt}

\usepackage{times}
\usepackage[edges]{forest}
\usepackage{adjustbox}

\usepackage[T1]{fontenc}  

\usepackage[utf8]{inputenc}

\usepackage{microtype}

\usepackage{inconsolata}
\usepackage{enumitem}

\usepackage{graphicx}
\usepackage{multirow} 

\title{
FormosanBench: Benchmarking Low-Resource Austronesian Languages in the Era of Large Language Models}

\author{Kaiying Kevin Lin$\ddagger$ \and Hsiyu Chen$\ddagger$ \and Haopeng Zhang$\dag$\\
{Institute of Linguistics, Academia Sinica$\ddagger$ \quad ALOHA Lab, University of Hawaii at Manoa$\dag$}\\
{\{limkhaiin, hsiyuchen\}@as.edu.tw \quad haopengz@hawaii.edu}
}

\begin{document}
\maketitle
\begin{abstract}
While large language models (LLMs) have demonstrated impressive performance across a wide range of natural language processing (NLP) tasks in high-resource languages, their capabilities in low-resource and minority languages remain significantly underexplored. Formosan languages—a subgroup of Austronesian languages spoken in Taiwan—are both linguistically rich and endangered, largely due to the sociolinguistic dominance of Mandarin. In this work, we introduce \textsc{FormosanBench}, the first benchmark for evaluating LLMs on low-resource Austronesian languages. It covers three endangered Formosan languages: Atayal, Amis, and Paiwan, across three core NLP tasks: machine translation, automatic speech recognition (ASR), and text summarization. We assess model performance in zero-shot, 10-shot, and fine-tuned settings using \textsc{FormosanBench}. Our results reveal a substantial performance gap between high-resource and Formosan languages. Existing LLMs consistently underperform across all tasks, with 10-shot learning and fine-tuning offering only limited improvements. These findings underscore the urgent need for more inclusive NLP technologies that can effectively support endangered and underrepresented languages. We release our datasets and code to facilitate future research in this direction : \url{https://anonymous.4open.science/r/FormosanBench-DB43/}.

\end{abstract}

\section{Introduction}

Large language models (LLMs) have achieved remarkable success across a wide range of natural language processing (NLP) tasks~\cite{hendrycks2020measuring,zhang2024systematic}, particularly for high-resource languages~\cite{magueresse2020low}. This success is largely driven by pretraining on massive datasets dominated by majority languages such as English, Mandarin, and Spanish~\cite{ruder2022square}. In contrast, the capabilities of LLMs in low-resource language settings remain significantly underexplored and underdeveloped ~\cite{joshi2025adaptingmultilingualllmslowresource}. Bridging this gap is critical for ensuring that NLP technologies are inclusive and beneficial to linguistically underrepresented communities.


Recent work has begun to address this disparity by developing benchmarks and datasets for low-resource languages. For example, \citet{ahia2024voicesunheardnlpresources} introduced datasets for Yorùbá and its dialects, while \citet{adelani2025irokobenchnewbenchmarkafrican} released IrokoBench, a multilingual benchmark for 17 typologically diverse African languages. These efforts underscore a growing recognition of the need for inclusive evaluation frameworks that go beyond the high-resource paradigm.

Formosan languages—a subgroup of Austronesian languages spoken in Taiwan—represent one such underrepresented group. With fewer than 200,000 speakers in total,\footnote{\url{https://en.wikipedia.org/wiki/Languages_of_Taiwan}} these languages are both linguistically rich and severely endangered. NLP development for Formosan languages faces unique challenges, particularly the scarcity of high-quality, annotated data~\cite{zheng-etal-2024-improving-low}. 



In this paper, we address these gaps by introducing \textsc{FormosanBench}, the first multi-task benchmark for Formosan languages, covering three endangered languages—Amis (ISO-ami), Atayal (ISO-tay), and Paiwan (ISO-pwn)—across three core NLP tasks: machine translation, automatic speech recognition (ASR), and text summarization. With \textsc{FormosanBench}, we systematically evaluate the zero-shot performance of state-of-the-art large language models (LLMs) and explore strategies for improving their performance through 10-shot in-context learning and small-scale fine-tuning. Experimental results reveal a significant performance gap for existing LLMs on Formosan languages, underscoring the need for further research. Our main contributions are as follows:
\begin{itemize}
    \vspace{-5pt}
    \item We release \textsc{FormosanBench}, the first multi-task benchmark for three major Formosan languages, supporting evaluation across machine translation, ASR, and summarization. We release our datasets and code to facilitate future research. 
    \vspace{-5pt}
    \item We conduct a comprehensive evaluation of multiple state-of-the-art LLMs on \textsc{FormosanBench} under zero-shot, 10-shot, and fine-tuning settings.
    \vspace{-5pt}
    \item Our results reveal a substantial performance gap on Formosan languages. While in-context learning and small-scale fine-tuning provide some improvements, the findings highlight the need for more effective and targeted adaptation methods for endangered and underrepresented languages.
    \vspace{-5pt}
\end{itemize}

\section{Background: Formosan languages}
The Austronesian language family is one of the world’s largest and most geographically dispersed, extending from Madagascar in the west to Hawaii and New Zealand in the east. Taiwan is widely considered the linguistic homeland of Austronesian, a hypothesis supported by the high degree of linguistic diversity found among its indigenous languages—collectively known as the Formosan languages. This diversity includes significant phonological, lexical, and syntactic variation, distinguishing Formosan languages from other Austronesian branches \citep{20fbbf33-f26e-3f27-bd7e-f1d8fac5c2e1, e23cd683-6d9e-337a-ab25-01f8e394d5e6, annurev:/content/journals/10.1146/annurev-linguistics-011718-012440}.

\begin{figure}[t]
    \centering
    \includegraphics[width=0.9\linewidth]{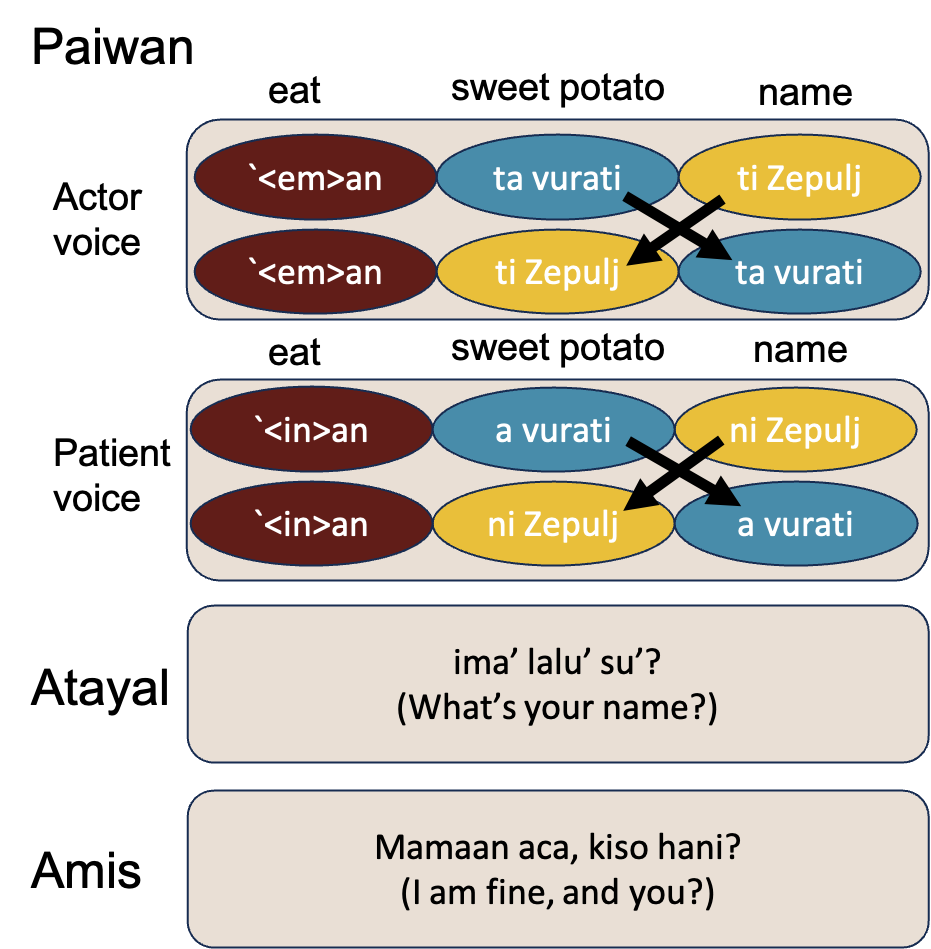}  
    \caption{Four patterns of a sentence 'Zepulj eats sweet potatoes' in Paiwan, and example sentences in Atayal and Amis.}
    \label{fig:formosan_sentence}
\end{figure}

Formosan languages pose considerable challenges for natural language processing (NLP), especially for large language models (LLMs). Like many other Austronesian languages, they feature verb-initial (VSO or VOS) syntax and a complex Voice system that is typologically rare. For instance, as illustrated in Figure~\ref{fig:formosan_sentence}, Paiwan allows multiple syntactic realizations of a sentence with essentially a similar meaning—e.g., ‘Zepulj eats sweet potatoes’—depending on how the verb is marked to highlight the roles of the noun phrases. The verb carries an agent-voice or patient-voice marker, and noun phrases change case marking accordingly and can switch positions  \citep{chang2018paiwan}. This flexibility, governed by a morphosyntactic voice system, is uncommon in the world's languages and adds complexity for modeling.

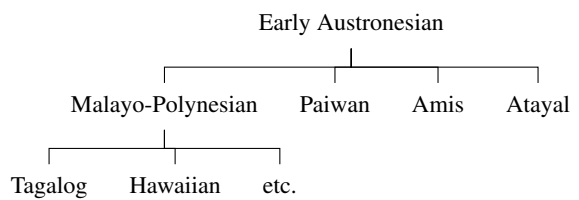
\begin{figure}[ht]
\centering
\begin{forest}
for tree={
    grow'=south,
    child anchor=north,
    parent anchor=south,
    edge path={
        \noexpand\path[\forestoption{edge}] (!u.parent anchor) -- +(0,-7pt) -| (.child anchor)\forestoption{edge label};
    },
    l sep=12pt,
    s sep=8pt,
    anchor=center,
    align=center,
    font=\footnotesize
}
[Early Austronesian 
  [Atayal]
  [Amis]
  [Paiwan]
  [Malayo-Polynesian
    [etc.] 
    [Hawaiian] 
    [Tagalog]
  ]
]
\end{forest}
\caption{Austronesian family tree.}
\label{fig:austronesian_tree}
\end{figure}

Further examples in Atayal and Amis (Figure~\ref{fig:formosan_sentence}) demonstrate similar typological features. These three languages—Atayal, Amis, and Paiwan—belong to early-diverging branches of the Austronesian family and are genealogically distant from each other as well as from more widely studied Malayo-Polynesian languages such as Hawaiian and Tagalog (see Figure~\ref{fig:austronesian_tree}). Their linguistic distinctiveness, combined with the scarcity of  pretraining corpora, makes them ideal candidates for studying LLM performance in low-resource and typologically diverse settings.

\begin{figure}[htbp]
    \centering
    \includegraphics[width=0.45\linewidth]{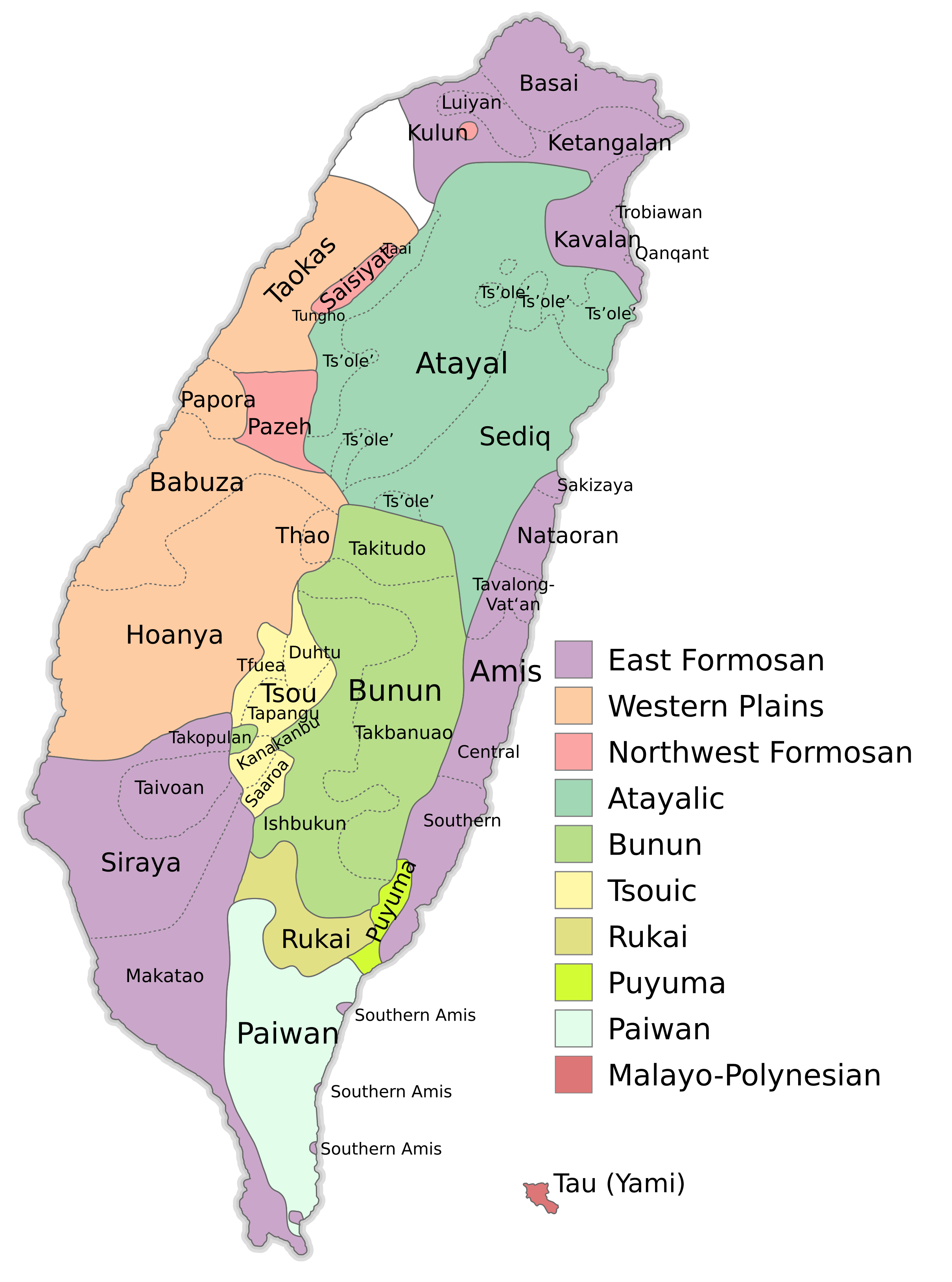}  
    \caption{16 Formosan languages in Taiwan}
    \label{fig:formosan_languages}
\end{figure}

\section{Related work}
Low-resource languages are those with limited linguistic data, tools, and representation in NLP research. An estimated 80–90\% of the world’s languages fall into this category \citep{hedderich2021surveyrecentapproachesnatural, joshi-etal-2020-state}.

\noindent\textbf{Low-resource NLP:} Recent advances in low-resource NLP have centered around three major areas: multilingual modeling, data preprocessing pipelines, and the development of benchmark datasets. Multilingual models such as those proposed by \citet{aharoni-etal-2019-massively} and \citet{conneau-etal-2020-unsupervised} demonstrate that training on large-scale corpora from over 100 languages can improve performance on tasks like machine translation, cross-lingual natural language inference, and general language understanding benchmarks.

To improve data quality, \citet{wenzek-etal-2020-ccnet} introduced CCNet, a scalable pipeline for cleaning noisy web-crawled text—an essential resource for expanding datasets in low-resource settings. In task-specific applications, \citet{mutsaddi-choudhary-2025-enhancing} enhanced plagiarism detection in the low-resource Indian language Marathi by combining BERT embeddings with TF-IDF, outperforming both methods used independently and achieving over 82\% accuracy.

More recently, researchers have introduced benchmark datasets to support a broader range of typologically diverse low-resource languages. \citet{ahia2024voicesunheardnlpresources} developed benchmark datasets for Yorùbá and its dialects, while \citet{shang2024atlaschatadaptinglargelanguage} introduced Atlas-Chat, a multi-task benchmark for Moroccan Arabic (Darija). They evaluated several state-of-the-art LLMs and demonstrated that few-shot and fine-tuned models significantly improved performance. These works highlight the growing focus on evaluation, benchmarking, and adaptation techniques for underrepresented languages in NLP.

\noindent\textbf{NLP in Formosan languages:} Despite growing interest in digital preservation, computational research on Formosan languages remains sparse. Among the few studies in this area, \citet{liao2023atayalicmt} proposed a Transformer-based machine translation system for three Atayalic languages—Atayal, Seediq, and Truku—using training data from online dictionaries and educational materials, and testing on manually curated book content. Similarly, \citet{zheng-etal-2024-improving-low} explored translation techniques across 16 Formosan languages by incorporating lexical resources and generating pseudo-parallel data through lexicon-based substitutions. While these approaches represent valuable steps forward, translation performance remains limited, reflecting both the complexity of Formosan languages and the scarcity of high-quality data.

Notably, existing work has focused narrowly on machine translation and conventional sequence models. No prior research has systematically evaluated the capabilities of state-of-the-art LLMs across a broader range of NLP tasks in these languages. This gap underscores the importance of developing comprehensive benchmarks and assessing the adaptability of LLMs in truly low-resource and typologically diverse language settings.

\begin{figure*}[t]
    \centering
    \includegraphics[width=\linewidth]{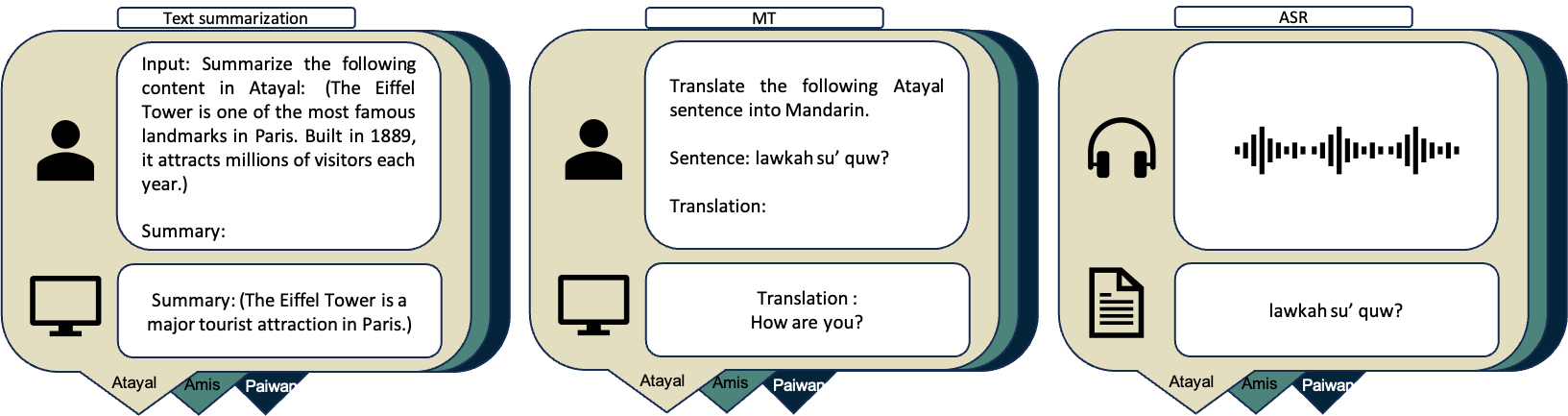} 
    \caption{Task Description for \textsc{FormosanBench} datasets. All the prompts were given in Mandarin, but we provide them in English for clarity. In the case of summarization, text appearing in the parentheses indicates content written in the Formosan language.}
    \label{fig:method}
\end{figure*}

\section{\textsc{FormosanBench}}
We present \textsc{FormosanBench}, the first benchmark to support three endangered Formosan languages—Atayal, Amis, and Paiwan—across three core NLP tasks: machine translation, automatic speech recognition, and text summarization. These languages were selected based on three criteria: 1) a combined speaker population exceeding 100,000; 2) the availability of existing digital resources; and 3) their linguistic diversity within the Austronesian language family. The choice of NLP tasks was guided primarily by data availability. Figure~\ref{fig:method} illustrates examples of each task.

\paragraph{Machine Translation} 
Our MT dataset comprises approximately 5,000 sentence pairs for each of three Formosan languages—Atayal, Amis, and Paiwan—paired with Mandarin. The data was extracted from the Taiwan Indigenous Languages E-Dictionary\footnote{\url{https://e-dictionary.ilrdf.org.tw/index.htm}} and underwent rigorous preprocessing as documented by ~\citet{formosanbank2024}\footnote{\url{https://github.com/FormosanBank/FormosanBank}
}.The preprocessing methodology included text normalization, punctuation correction, removal of encoding artifacts, and XML structure validation to ensure data quality across sentence-level units. We extracted the Formosan sentences and their Mandarin counterparts from the preprocessed XML files to form the final parallel corpora.


\paragraph{ASR}
We collected the ASR data from FormosanBank. The data were sourced from the \textit{Klokā Digital Platform for Indigenous Languages}\footnote{\url{https://web.klokah.tw/}}, which hosts digitized textbooks for 16 Formosan languages aimed at learners of all ages. Each textbook section contains written text in a Formosan language along with corresponding audio recordings, produced by native speakers reading the content aloud. The materials cover typical language-learning themes—such as reading and writing, picture books, daily conversations, short texts, situational language use, illustrated stories, and cultural topics—offering a range of speech styles, vocabulary, and sentence complexity grounded in culturally specific content. For this study, we selected the dialects Coastal Amis, Squliq Atayal, and Central Paiwan respectively, due to their relative representativeness and greater accessibility within available resources. The final compiled datasets consist of paired sentence-level audio recordings with their corresponding transcriptions, manually verified for alignment accuracy.

\paragraph{Summarization}
Following ~\citet{liu2018generatingwikipediasummarizinglong}, we collected Wikipedia articles written in Formosan languages. These articles often focus on culturally significant topics such as indigenous traditions, historical narratives, and regional geography~\cite{yuan2024domainsum}. Notably, the articles follow a relatively consistent structure: an initial introductory section summarizing the topic, followed by a series of subsections with more detailed expository content. This natural hierarchy of information lends itself well to the single-document summarization task, where the introduction can serve as a gold-standard summary and the remainder of the article as the input document ~\cite{ta2022wikideswikipediabaseddatasetgenerating,zhang2022improving}.
Based on this structure, we curated the dataset by pairing each article’s introductory section with the corresponding main body content. We manually reviewed the articles to ensure coherence between the summary and body, and excluded entries that lacked sufficient structure or were too short to support meaningful summarization.


\begin{table}[t]
\centering
\small
\renewcommand{\arraystretch}{1.3}
\begin{tabular}{p{1.6cm}|
  >{\centering\arraybackslash}p{1.6cm}
  >{\centering\arraybackslash}p{1.6cm}
  >{\centering\arraybackslash}p{1.6cm}}
\hline
\textbf{Language} & \textbf{Amis} & \textbf{Atayal} & \textbf{Paiwan} \\
\hline
\multicolumn{4}{c}{\textbf{Machine Translation (MT)}} \\
\hline
Datapoints & 3,259 / 543 / 1,631 & 3,832 / 638 / 1,918 & 3,216 / 536 / 1,609 \\
Total Words & 38,946 & 38,946 & 36,601 \\
\hline
\multicolumn{4}{c}{\textbf{Automatic Speech Recognition (ASR)}} \\
\hline
Datapoints & 2,567 / 425 / 1,288 & 2,845 / 470 / 1,429 & 2,054 / 339 / 1,050 \\
Total Words & 55,478 & 54,781 & 29,677 \\
\hline
\multicolumn{4}{c}{\textbf{Summarization}} \\
\hline
Datapoints & 77 / 21 / 41 & 672 / 116 / 299 & 135 / 27 / 69 \\
Sum Length& 10,438 & 18,983 & 68,537 \\
Doc Length & 90,477 & 75,303 & 198,316 \\
\hline
\end{tabular}
\caption{Dataset statistics and splits (Train/Val/Test) for MT, ASR, and summarization tasks across the three Formosan languages. Word counts are in Formosan languages.}
\vspace{-4pt}
\label{tab:dataset_info}
\end{table}

\paragraph{Quality Control}
To ensure the reliability and usefulness of our datasets, we applied task-specific quality control procedures. For the MT dataset, we apply rule-based filter to remove duplicate sentence pairs to avoid redundancy and evaluation bias.

For the ASR dataset, we reviewed the XML metadata associated with the selected Formosan languages from FormosanBank. We excluded entries that met any of the following criteria: 1) instances containing only a single word, 2) entries lacking corresponding audio file links sourced from the associated textbooks, and 3) duplicate entries.

The summarization datasets underwent additional filtering to address quality issues in Wikipedia articles written in Formosan languages. Due to content sparsity, many entries contained only introductory paragraphs without substantial body text. We implemented a two-stage filtering process: first removing datapoints where article bodies consisted solely of section titles without accompanying content, then excluding entries where the input text (article body) was less than 1.5 times the length of the summary (introduction). This ensured that remaining samples provided meaningful training and evaluation material with sufficient information density for effective summarization model development.

\paragraph{Benchmark Statistics}
Detailed statistics for each task in \textsc{FormosanBench} are presented in Table~\ref{tab:dataset_info}. For each Formosan language, the dataset was divided into training, validation, and test splits using a 70:10:20 ratio. This consistent split facilitates standardized evaluation across all tasks and languages. Among the three languages, Atayal provides the largest number of data points and word tokens across all tasks, supporting broader coverage and diversity. In contrast, Paiwan offers the smallest datasets, with fewer examples and a lower total word count, particularly in the ASR and summarization tasks. These differences reflect resource availability and highlight the varying levels of vocabulary richness and content complexity among the languages.

\section{Experiments}

\subsection{LLMs used for evaluation}
\paragraph{Translation and Summarization}

We selected a combination of open-source and proprietary state-of-the-art LLMs, with a focus on multilingual capabilities and their effectiveness in low-resource language settings:

\begin{itemize}
    \vspace{-4pt}
    \item \textbf{LLaMA 3.1}~\citep{meta2024llama3}: An autoregressive transformer developed by Meta, trained on a diverse multilingual and code corpus. It provides a strong baseline for general-purpose multilingual tasks.  
    \vspace{-6pt}
    \item \textbf{Gemma-1.1}~\citep{gemmateam2024gemmaopenmodelsbased}: A compact, instruction-tuned model from Google, optimized for efficient inference and strong zero-shot reasoning abilities. 
    \vspace{-6pt}
    \item \textbf{Mistral-7B v0.3}~\citep{jiang2023mistral7b}: A 7-billion-parameter model that incorporates sliding window attention and grouped-query attention, enabling efficient multilingual inference.
    \vspace{-6pt}
    \item \textbf{NLLB-200}~\citep{nllbteam2022languageleftbehindscaling}: Meta’s encoder-decoder transformer specifically designed for low-resource machine translation, offering a complementary architecture to decoder-only LLMs.
    \vspace{-6pt}
    \item \textbf{GPT-4o}~\citep{openai2024gpt4o}: OpenAI’s latest proprietary model, known for its advanced zero-shot and multilingual capabilities. Prior studies (e.g., \citealp{adelani2025irokobenchnewbenchmarkafrican}) suggest that such proprietary models often outperform open-source ones in low-resource settings, and we include it as an upper-bound baseline.
    \vspace{-4pt}
\end{itemize}

We used consistent prompting strategies for both MT and summarization tasks across all models. Example prompts are provided in Appendix~\ref{sec:appendix}.

\paragraph{Automatic Speech Recognition}

For the ASR task, we evaluated a set of state-of-the-art models with demonstrated performance in multilingual and low-resource contexts~\cite{ahia2024voicesunheardnlpresources}:

\begin{itemize}
    \vspace{-4pt}
    \item \textbf{Whisper}~\citep{radford2022robustspeechrecognitionlargescale}: An encoder-decoder transformer trained on 680,000 hours of multilingual and multitask supervised data. It is well-regarded for robustness in noisy and low-resource scenarios. 
    \vspace{-6pt}
    \item \textbf{SeamlessM4T}~\citep{communication2023seamlessm4tmassivelymultilingual}: A unified, multilingual, and multimodal system developed by Meta, capable of handling speech and text in a single encoder-decoder framework for ASR, translation, and generation tasks.
    \vspace{-6pt}
    \item \textbf{MMS-1b-all}~\citep{pratap2023mms}: Part of Meta's Massively Multilingual Speech project, this model uses a conformer-based encoder (built on \textsc{Wave2Vec 2.0}~\citep{baevski2020wav2vec20frameworkselfsupervised}) and supports ASR and language ID for over 1,000 languages.
    \vspace{-4pt}
\end{itemize}

\subsection{Implementation details}

For the MT task, we evaluated performance in both directions: from Formosan languages to Mandarin and from Mandarin to Formosan languages. We applied 10-shot in-context learning and full fine-tuning to two models: Mistral-7B v0.3 and NLLB. For the summarization task, we performed 10-shot learning with GPT-4o, Mistral-7B v0.3, and LLaMA 3.1 8B; the latter two models were also fine-tuned on task-specific data. All fine-tuning was performed using the Hugging Face Trainer with a learning rate of 0.001, batch size of 4, and 20 epochs. The checkpoint with the lowest evaluation loss was selected as the final model.

For ASR, the MMS model requires explicit specification of the transcription language. We selected Amis as the target language, since Atayal and Paiwan were not included in MMS’s pretraining corpus. In contrast, Whisper and Seamless support automatic language detection and were used without manual language specification. No prompts were required for the ASR task. Among these, Whisper was the only model fine-tuned on Formosan language data, due to its robust open-source implementation, wide community adoption, and the availability of streamlined tools and APIs that support task-specific adaptation. Fine-tuning was conducted with a batch size of 16, learning rate of 0.0001, and a maximum of 5000 training steps. The final model was selected based on the lowest word error rate (WER) on the evaluation set.

All experiments were conducted within the Academia Sinica AI development container environment on NVIDIA V100 GPUs and PyTorch 2.7.0a0.

\subsection{Evaluation metrics}
We report BLEU scores \citep{10.3115/1073083.1073135} for the machine translation (MT) task, while the ASR and summarization tasks are evaluated using word error rate (WER) and ROUGE scores \citep{lin-2004-rouge}, respectively. BLEU measures n-gram precision (typically up to 4-grams) between model outputs and reference translations, with a brevity penalty to discourage overly short outputs. WER quantifies the proportion of word-level errors (insertions, deletions, substitutions) in transcribed speech, while ROUGE evaluates summary quality based on overlapping units such as bigrams (ROUGE-2) or longest common subsequences (ROUGE-L) between generated and reference summaries. For GPT-4o outputs, we additionally collect human judgments to qualitatively assess fluency and content relevance.
\section{Results and Discussion}

\subsection{MT Results}

\begin{table*}[t]
\centering
\scriptsize
\renewcommand{\arraystretch}{1.2}
\begin{tabular}{l|ccc|ccc}
\hline
\multirow{2}{*}{} 
  & \multicolumn{3}{c|}{\textbf{Formosan $\rightarrow$ Mandarin}} 
  & \multicolumn{3}{c}{\textbf{Mandarin $\rightarrow$ Formosan}} \\
  & \textbf{Amis} & \textbf{Atayal} & \textbf{Paiwan} 
  & \textbf{Amis} & \textbf{Atayal} & \textbf{Paiwan} \\
\hline
\multicolumn{7}{c}{\textbf{Zero-shot Models}} \\
\hline
Mistral-7B-Instruct     & 0.0000 & 0.0000 & 0.0000 & 0.0000 & \textbf{0.0010} & 0.0002 \\
Gemma-1.1-7B-it         & 0.0000 & 0.0000 & 0.0000 & 0.0000 & 0.0005 & 0.0019 \\
LLaMA-3.1-8B            & 0.0000 & 0.0000 & 0.0000 & 0.0000 & 0.0001 & 0.0000 \\
LLaMA-3.1-70B           & 0.0000 & 0.0000 & 0.0000 & 0.0000 & 0.0006 & 0.0005 \\
GPT-4o-mini             & 0.0000 & 0.0000 & 0.0000 & 0.0000 & 0.0001 & \textbf{0.0036} \\
NLLB                   & 0.0000 & 0.0000 & 0.0000 & 0.0000 & 0.0000 & 0.0000 \\
\hline
\multicolumn{7}{c}{\textbf{10-shot Models}} \\
\hline
Mistral-10-shot         & 0.0000 & 0.0000 & 0.0035 & 0.0016 & 0.0000 & 0.0041 \\
\hline
\multicolumn{7}{c}{\textbf{Fine-tuned Models}} \\
\hline
Mistral-Finetuned       & 0.0000 & 0.0000 & 0.0019 & 0.0159 & 0.0208 & 0.0002 \\
NLLB-Finetuned          & 0.0000 & \textbf{0.0048} & \textbf{0.0759} & \textbf{0.1506} & \textbf{0.1137} & \textbf{0.0636} \\
\hline
\end{tabular}
\caption{BLEU scores $\uparrow$ for translation between Formosan languages and Mandarin. Left: Formosan $\rightarrow$ Mandarin. Right: Mandarin $\rightarrow$ Formosan. The best score for each language in each condition is in bold.}.
\vspace{-4pt}
\label{tab:bleu_score}
\end{table*}

\begin{figure}[t]
    \centering
    \includegraphics[width=\linewidth]{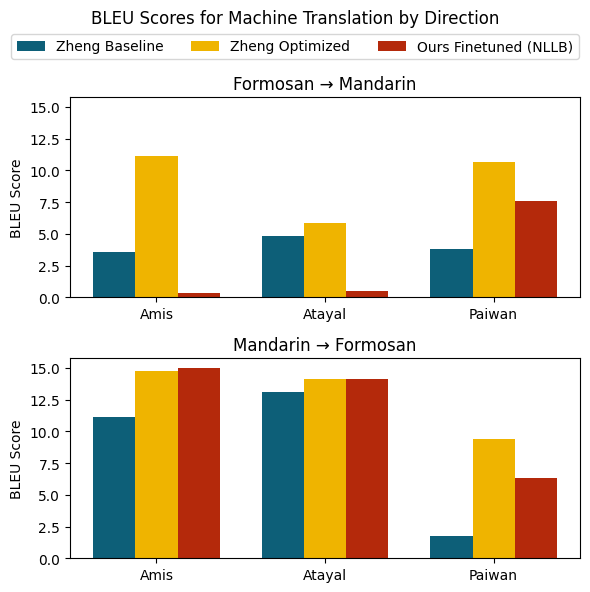}  
    \caption{BLEU scores $\uparrow$ compared with Zheng et al. (2024)}
    \label{fig:bleu_compared_with_zheng}
    \vspace{-10pt}
\end{figure}

\noindent\textbf{Zero-shot Setting}. As shown in Table ~\ref{tab:bleu_score}, across all non-fine-tuned models—including GPT-4o, LLaMA-3.1 (8B and 70B), Gemma-1.1, and Mistral-7B-instruct variants—the BLEU scores remained consistently low, often at or extremely close to zero, and even a proprietary model like GPT-4o did not show better performance. In the Formosan-to-Mandarin direction, all models scored nearly 0 BLEU across all three languages, indicating a near-total failure to understand and translate from these low-resource languages. In the reverse direction, Mandarin-to-Formosan, performance was only marginally higher: for example, the highest non-fine-tuned BLEU scores were just 0.0019 for Paiwan (Gemma-1.1-7B-it), 0.0010 for Atayal (Mistral-7B-Instruct), and 0.0036 for Paiwan (GPT-4o-mini). These small improvements suggest that the models have stronger generative capacity when translating into Formosan languages from Mandarin input, likely due to their familiarity with Mandarin and surface-level token overlap. However, the overall low BLEU scores still reflect minimal understanding of Formosan linguistic structure or vocabulary.

\noindent\textbf{10-shot and Fine-tuning Settings}. We observed that BLEU scores for the Mistral models remained nearly zero even after adaptation. In contrast, NLLB showed slight improvements, likely due to its MT-oriented architecture and pretraining on a multilingual corpus that includes Austronesian languages such as Tagalog and Indonesian. Nevertheless, the gains remain minimal, highlighting the challenges of generalizing from these languages to Formosan languages. Our best results in each section are compared with those reported in \citet{zheng-etal-2024-improving-low}, who trained models from scratch. Figure~\ref{fig:bleu_compared_with_zheng} shows our best training results from NLLB models, compared with \citet{zheng-etal-2024-improving-low}. Our results remain generally low, and adaptation contributes little to performance improvement, even when compared to \citet{zheng-etal-2024-improving-low}’s already modest results. The only exceptions from NLLB fine-tuning with non-parallel Formosan data, which marginally exceeded \citet{zheng-etal-2024-improving-low}’s baseline, suggest that future improvements may require language-specific adaptation strategies. Overall, the limited amount of training data remains a key bottleneck—current models, with millions to billions of parameters, require far more data than what is currently available for Formosan languages.

\subsection{ASR Results}
\noindent\textbf{Zero-shot Setting}. Table~\ref{tab:wer_formosan} presents the word error rate (WER) results for Amis, Atayal, and Paiwan. Among these models, MMS consistently achieved the lowest WER across all three languages, while Whisper Base and Seamless exhibited substantially higher error rates. Notably, Amis—explicitly included in MMS’s pretraining data and specified for our experiments—showed the strongest performance, as expected. While MMS’s familiarity with Amis may have provided marginal benefits to its performance on Atayal and Paiwan, these gains were limited. As discussed earlier in the paper, Atayal and Paiwan exhibit linguistic differences from Amis, which likely constrained the model’s ability to generalize effectively across these languages. The overall stronger performance of MMS can also be attributed to its large-scale multilingual training and its architecture specifically optimized for low-resource language settings. These findings are consistent with prior research by \citet{ahia2024voicesunheardnlpresources}, which similarly reported that MMS outperforms general-purpose ASR models in under-resourced language settings.

\begin{figure}[t]
    \centering
    \includegraphics[width=\linewidth]{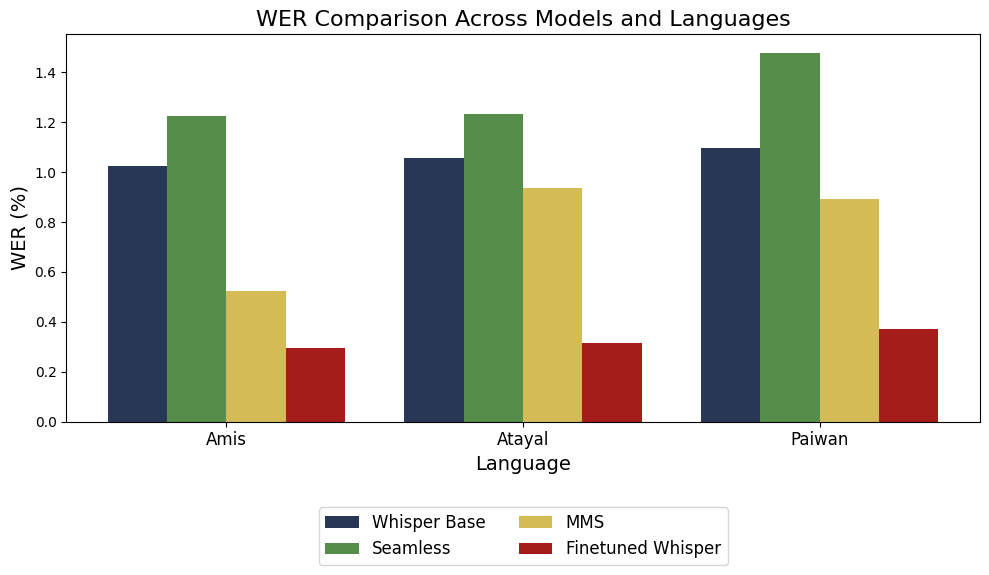} 
    \caption{WER $\downarrow$ comparison across models and languages}
    \label{fig:wer_comparison}
 \vspace{-20pt}\end{figure}

\noindent\textbf{10-shot and Fine-tuning Settings}. As shown in Figure \ref{fig:wer_comparison}, the fine-tuned Whisper model shows a notable reduction in WER, achieving scores around 0.3. This improvement may be attributed to Whisper’s extensive pretraining on hundreds of thousands of hours of multilingual and multitask supervised data collected from the web. While this marks a significant improvement, further adaptation may still be necessary to reach a level suitable for practical deployment.

\begin{table}[t]
\centering
\small
\renewcommand{\arraystretch}{1.2}
\begin{tabular}{lccc}
\hline
 & \textbf{Amis} & \textbf{Atayal} & \textbf{Paiwan} \\
\hline
\multicolumn{4}{c}{\textbf{Zero-shot Models}} \\
\hline
Whisper Base        & 1.026 & 1.057 & 1.098 \\
Seamless            & 1.225 & 1.235 & 1.479 \\
MMS                 & \textbf{0.522} & \textbf{0.937} & \textbf{0.893} \\
\hline
\multicolumn{4}{c}{\textbf{Fine-tuned Model}} \\
\hline
Finetuned Whisper   & \textbf{0.296} & \textbf{0.315} & \textbf{0.373} \\
\hline
\end{tabular}
\caption{WER (Word Error Rate) $\downarrow$ comparison across ASR models for Amis, Atayal, and Paiwan.}
\vspace{-4pt}
\label{tab:wer_formosan}
\end{table}

\subsection{Text summarization Results}
\noindent\textbf{Zero-shot Setting}. Table~\ref{tab:rouge_reusults} shows that performance on the summarization tasks was generally low, with ROUGE-2 and ROUGE-L scores typically falling below 20. The generated summaries often included Mandarin words with only a few Formosan words, indicating that the models struggled to understand and generate content in the target languages. Despite low overall comprehension in both the translation and summarization tasks, ROUGE scores appeared relatively higher than BLEU scores. This discrepancy likely stems from ROUGE’s recall-oriented nature: when models produced lengthy but semantically irrelevant summaries, the chance of word overlap with the reference—particularly for frequent or proper nouns—increased. Even minimal overlap can contribute positively to the ROUGE score. In contrast, BLEU applies stricter penalties for mismatches, especially in terms of word order and precision, which helps explain the lower BLEU scores.

\noindent\textbf{10-shot and Fine-tuning Settings}. Model adaptation did not yield consistent improvements in ROUGE scores for Mistral and LLaMA; in fact, scores declined across most languages and models. This pattern likely reflects the models’ limited capacity to generate meaningful summaries in low-resource settings. ROUGE scores below 20 are generally considered unreliable as indicators of semantic quality, and fluctuations at this range are more likely attributable to noise than substantive gains. Prior to adaptation, some model outputs occasionally included key terms—such as named entities or location markers—that matched the reference summaries, likely due to memorization from pretraining.

After adaptation, the generated summaries appeared more fluent in Formosan languages but often lacked semantic coherence. Notably, the frequency of overlapping key terms with reference summaries decreased, likely contributing to the observed decline in ROUGE scores. An exception was GPT-4o, which exhibited substantial improvements in the summarization task for Atayal and Amis following 10-shot learning. Human evaluation indicated that many inputs involved descriptions of locations or tribes, typically including details about ethnic composition and geographic context. When few-shot examples featured similar content, GPT-4o effectively reproduced the structural and lexical patterns of the input examples, generating summaries with comparable tone and format. However, the model frequently introduced factual inaccuracies—such as incorrect ethnic group counts—and outputs on other topics, particularly those describing countries and locations, with significant grammatical errors. These findings suggest that the model's improvements stem more from surface-level pattern replication than from genuine semantic understanding.

In summary, most models struggled to perform the summarization task effectively. While GPT-4o showed notable improvements for specific languages under 10-shot learning, its gains did not generalize broadly. The overall performance gap between low-resource Formosan languages and high-resource counterparts underscores the need for more targeted strategies to improve LLM performance in typologically diverse, underrepresented languages.

\begin{table}[t]
\centering
\small
\renewcommand{\arraystretch}{1.2}
\begin{tabular}{lccc}
\hline
 & \textbf{Amis} & \textbf{Atayal} & \textbf{Paiwan} \\
\hline
\multicolumn{4}{c}{\textbf{Zero-shot Models}} \\
\hline
Mistral   & \textbf{19.74}/\textbf{24.75} & \textbf{11.10}/\textbf{25.34} & 13.20/18.66 \\
Gemma      & 1.32/7.68   & 1.76/7.58   & 2.91/10.32 \\
LLaMA              & 4.94/14.46  & 2.46/12.86  & 5.05/17.44 \\
GPT-4o                & 5.03/14.52  & 2.91/13.03  & \textbf{6.28}/\textbf{19.41} \\
\hline
\multicolumn{4}{c}{\textbf{10-shot Models}} \\
\hline
Mistral      & 0.94/7.65 & 0.52/6.02   & 0.24/8.11 \\
GPT-4o & \textbf{19.6}/\textbf{30.02} & \textbf{38.36}/\textbf{50.41} & \textbf{6.48}/\textbf{20.05} \\
LLaMA        & 0.59/7.15    & 0.52/5.11    & 0.27/6.53 \\
\hline
\multicolumn{4}{c}{\textbf{Fine-tuned Models}} \\
\hline
Mistral    & 3.87/10.20   & \textbf{11.02}/\textbf{19.57} & 0.50/\textbf{9.14} \\
LLaMA      & \textbf{4.60}/\textbf{13.32} & 5.81/10.39 & \textbf{0.97}/7.93 \\
\hline
\end{tabular}
\caption{ROUGE-2/ROUGE-L scores $\uparrow$ for Amis, Atayal, and Paiwan across model types. All values are scaled to percentages for interpretability.}
\vspace{-4pt}
\label{tab:rouge_reusults}
\end{table}

\section{Conclusion}
In this paper, we present evaluation results from several large language models (LLMs) across three Formosan languages. Our findings demonstrate that low-resource languages, such as those in the Formosan family, remain a significant blind spot for current state-of-the-art LLMs. The models perform poorly across all three evaluated NLP tasks—machine translation (MT), automatic speech recognition (ASR), and text summarization. In particular, results from MT and summarization indicate that the models struggle to exhibit meaningful understanding of the target languages, often producing outputs that are entirely unrelated to the input. While ASR performance is relatively better, it still falls short of practical usability. Notably, our adaptation efforts led to a substantial reduction in word error rate (WER) for ASR and a notable increase in ROUGE scores for GPT-4o in summarization. However, improvements in BLEU scores for MT were marginal, and most models showed a decline in ROUGE scores on the summarization task. Overall, these benchmarking results underscore the urgent need for targeted research and model development tailored to low-resource languages such as those in the Formosan family.

\section*{Limitations}
We present results for only a limited number of Formosan languages, and incorporating a broader range of NLP tasks would help build a more comprehensive understanding of low-resource language performance. However, expanding to additional tasks often demands substantial human effort for data annotation and validation, which is both time- and cost-consuming. These constraints significantly limit the scope of the current study. Future work, if supported by greater time and funding, may expand the scale of research in these directions.


\bibliography{custom}
\newpage
\appendix

\section{Prompts used in the experiments}
The detailed prompts used in all experiments are presented in Table~\ref{tab:prompt_table}.
\label{sec:appendix}
\begin{table*}[ht]
\centering
\begin{tabular}{ll}
\hline
\textbf{Task} & \textbf{Prompt Template} \\
\hline
MT (Formosan $\rightarrow$ Mandarin) & \verb|Translate the following {language} sentence into Mandarin.|\\
    &\verb|{Sentence}|\\
    & \verb|Transaltion: {Sentence}| \\
\hline
MT (Mandarin $\rightarrow$ Formosan) & \verb|Translate the following Mandarin sentence into {language} |\\
    &\verb|{Sentence}|\\
    & \verb|Transaltion: {Sentence}| \\
\hline
ASR & \verb|N/A| \\
\hline
Summarization & \verb|You are an expert in the {language} language.| \\
         & \verb|Summarize the following content in {language}.| \\
         & \verb|Your output must be in {language} only.|\\
         & \verb|Do not use Chinese or English.| \\
         & \verb|Only provide the summary: {text}| \\
\hline
\end{tabular}
\caption{Prompts used in our experiment (in Mandarin)}
\label{tab:prompt_table}
\end{table*}

\end{document}